\documentclass{article}

    \usepackage[preprint,nonatbib]{neurips_data_2024}

\usepackage[dvipsnames,table]{xcolor}

\usepackage[utf8]{inputenc} 
\usepackage[T1]{fontenc}    
\usepackage{hyperref}       
\usepackage{url}            
\usepackage{booktabs}       
\usepackage{amsfonts}       
\usepackage{nicefrac}       
\usepackage{microtype}      
\usepackage{xcolor}         

\usepackage{graphicx}
\usepackage{amsmath,amssymb} 

\usepackage{csquotes}
\usepackage{pifont}
\usepackage{multirow}
\usepackage{multicol}
\usepackage{appendix}

\title{DefAn: Definitive Answer Dataset for LLMs Hallucination Evaluation}

\author{%
  A B M Ashikur Rahman\thanks{Corresponding author} \\
  ICS Department\\ KFUPM\\
  Dhahran, KSA - 31261 \\
  \texttt{g202204800@kfupm.edu.sa} \\
  \And
  Saeed Anwar\\
  ICS Department, KFUPM\\
  JRCAI, SDAIA-KFUPM\\
  Dhahran, KSA - 31261 \\
  \texttt{saeed.anwar@kfupm.edu.sa} \\
  \And
  Muhammad Usman\\
  ICS Department, KFUPM\\
  JRCAI, SDAIA-KFUPM\\
  Dhahran, KSA - 31261 \\
  \texttt{muhammad.usman@kfupm.edu.sa} \\
  \And
  Ajmal Mian\\
  The University of Western Australia\\
  Crawley, Western Australia \\
  \texttt{ajmal.mian@uwa.edu.au} \\
}

\begin{document}

\maketitle

\begin{abstract}

Large Language Models (LLMs) have demonstrated remarkable capabilities, revolutionizing the integration of AI in daily life applications. However, they are prone to hallucinations, generating claims that contradict established facts, deviating from prompts, and producing inconsistent responses when the same prompt is presented multiple times. Addressing these issues is challenging due to the lack of comprehensive and easily assessable benchmark datasets. Most existing datasets are small and rely on multiple-choice questions, which are inadequate for evaluating the generative prowess of LLMs. To measure hallucination in LLMs, this paper introduces a comprehensive benchmark dataset comprising over 75,000 prompts across eight domains. These prompts are designed to elicit definitive, concise, and informative answers. The dataset is divided into two segments: one publicly available for testing and assessing LLM performance and a hidden segment for benchmarking various LLMs. In our experiments, we tested six LLMs—GPT-3.5, LLama 2, LLama 3, Gemini, Mixtral, and Zephyr—revealing that overall factual hallucination ranges from 59\% to 82\% on the public dataset and 57\% to 76\% in the hidden benchmark. Prompt misalignment hallucination ranges from 6\% to 95\% in the public dataset and 17\% to 94\% in the hidden counterpart. Average consistency ranges from 21\% to 61\% and 22\% to 63\%, respectively. Domain-wise analysis shows that LLM performance significantly deteriorates when asked for specific numeric information while performing moderately with person, location, and date queries. Our dataset demonstrates its efficacy and serves as a comprehensive benchmark for LLM performance evaluation. Our dataset and LLMs responses are available at  \href{https://github.com/ashikiut/DefAn}{https://github.com/ashikiut/DefAn}. 
\end{abstract}

\section{Introduction}
The domain of Generative artificial intelligence (AI) has witnessed a paradigm shift with the emergence of Large Language Models (LLMs). These powerful AI models, capable of processing and generating human-like text, have become ubiquitous across diverse applications. From facilitating seamless machine translation and engaging chatbot interactions to composing creative content and generating code, LLMs have demonstrably revolutionized numerous fields~\cite{Humza2023LLMs}. However, their immense potential is marred by a critical challenge–Hallucinations~\cite{rawte2023survey}.

Hallucination is characterized as the LLM-generated response that lacks coherence or deviates from the original source material~\cite{ji2023survey}. In other words, Hallucination generates a response that deviates from the user prompt or previously generated context~\cite{adlakha2023evaluating} or contradicts established fact~\cite{muhlgay2023generating}. These hallucinations manifest in various forms, ranging from demonstrably false information to content that differs significantly from the context of the prompt~\cite{zhang2023siren}. The ability of LLMs to generate such misleading information poses a significant threat to their trustworthiness, particularly in contexts where factual accuracy and adherence to prompts are critical.

\begin{figure}[tp]
\centering
\begin{tabular}{c|c}
\includegraphics[width= .49 \linewidth]{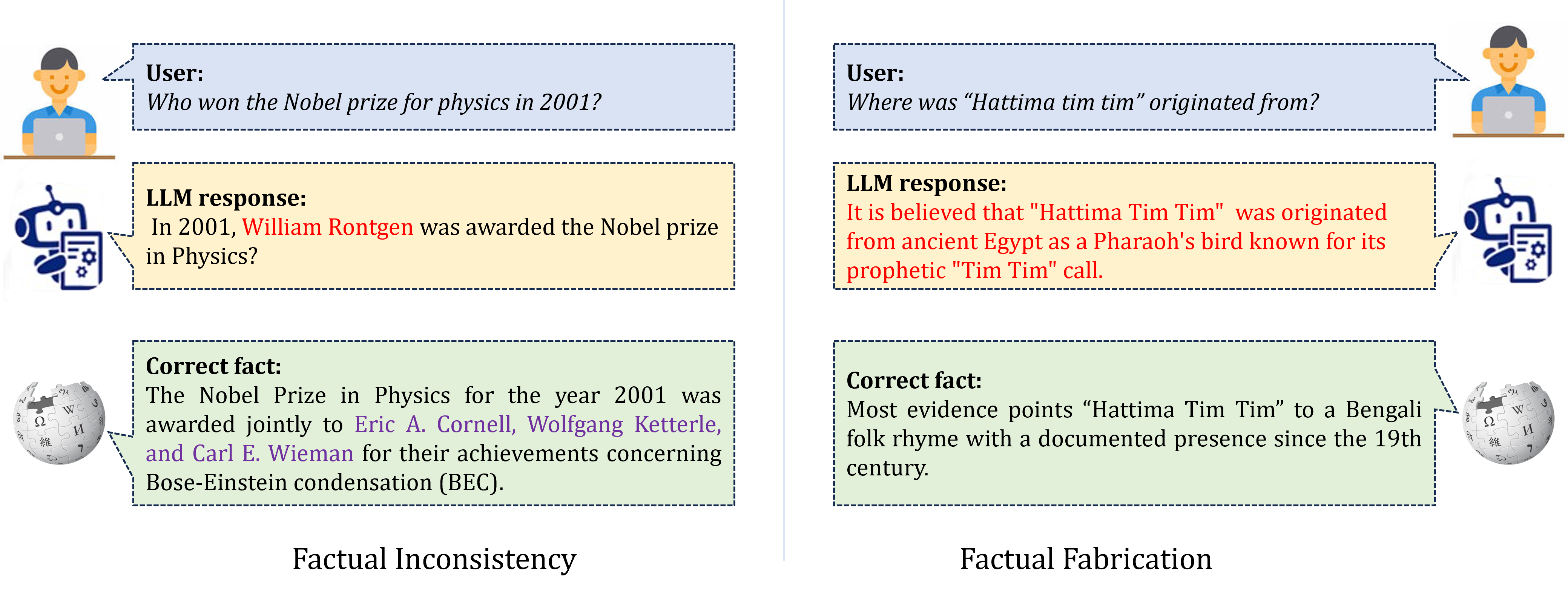}
&
\includegraphics[width= .49 \linewidth]{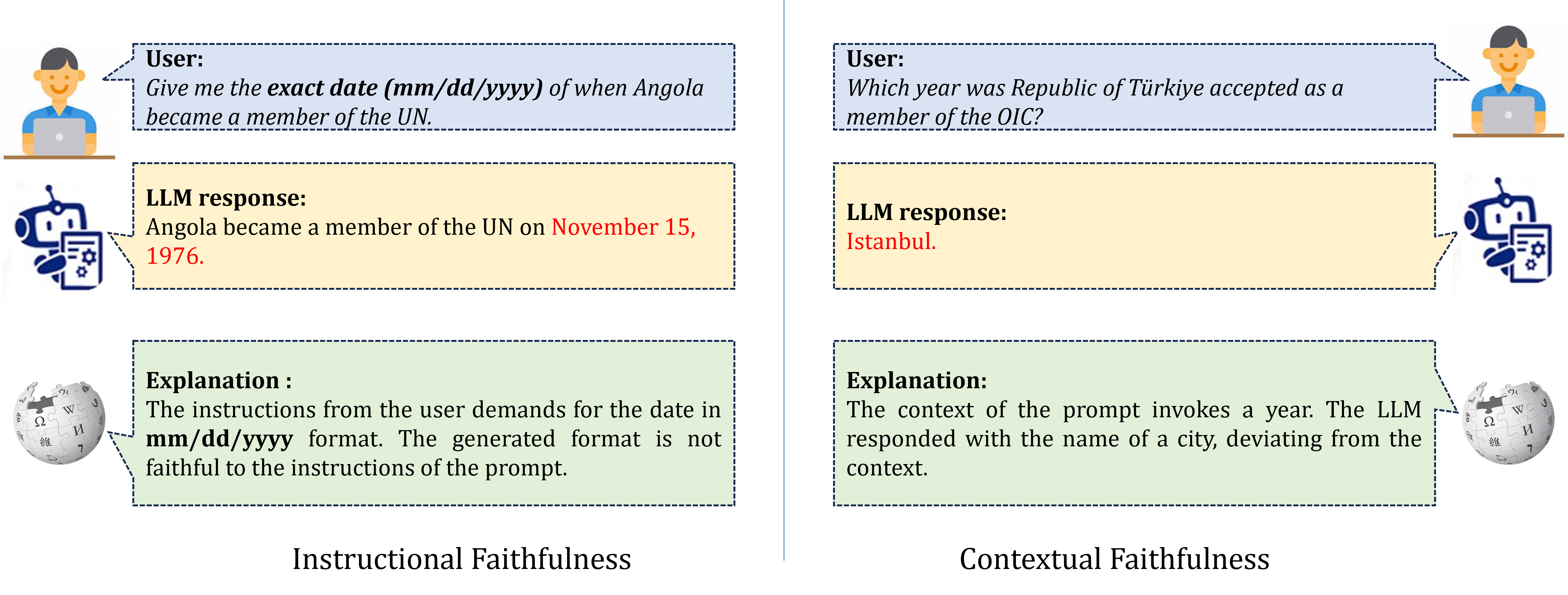}\\
a) & b) \\
\end{tabular}
\caption{Comparison between different types of hallucinations. a) Fact Contradicting Hallucinations and b) Prompt Misalignment Hallucinations. Best viewed on a zoomed-in screen.}
\label{fig:halu_types}
\end{figure}

Hallucinations can be grouped from different viewpoints. One such perspective broadly categorizes the hallucination into two main types: contradiction to fact and prompt misalignment. Factual hallucinations address the truthfulness of the generated content. They can be further divided into factual inconsistency, where the information contradicts existing facts, and factual fabrication, where entirely new, unverified information is created, as shown in Figure~\ref{fig:halu_types}(a). Prompt Misalignment, on the other hand, focuses on the deviation from the intent and context of the prompt. These can be instructional hallucinations, in which the LLM ignores specific instructions within the prompt, or contextual hallucinations, in which the generated response deviates from the prompt's overall theme or style. Examples are provided in Figure~\ref{fig:halu_types}(b)

    

Detecting and mitigating hallucinations remains a complex task in LLM research. Evaluation benchmarks play a significant role in comprehending an LLM's hallucination level. These benchmarks function as essential tools for assessing the trustworthiness of LLMs by providing a structured framework for evaluating their susceptibility to generating hallucinations~\cite{huang2023survey}. While commendable efforts have led to the development of benchmarks like FELM~\cite{zhao2024felm}, HaluEval~\cite{li2023halueval}, and HaluEval-Wild~\cite{HaluEval-Wild}, the current landscape of LLM evaluation datasets remains inadequate. One fundamental limitation is that most existing benchmarks have a narrow focus. Many prioritize either factual hallucinations or prompt misalignment, neglecting the multifaceted nature of LLM hallucinations. Additionally, relying on metrics derived from LLM-judge (a performance assessment model) raises concerns about inherent biases and potential inaccuracies within these metrics. Human evaluation, while desirable for achieving the highest level of accuracy, quickly becomes impractical when dealing with large datasets.

We propose a novel approach to address the limitations mentioned above by introducing a large-scale benchmark dataset, meticulously crafted to comprehensively evaluate three critical aspects of LLM performance:
\begin{itemize}
\itemsep0pt    \item Factual Accuracy: This facet assesses the LLM's ability to generate information grounded in verifiable reality.
    \item Faithfulness to the Prompt: Here, the focus shifts to evaluating how well the LLM adheres to the intent and style of the provided prompt.
    \item Consistency of Generated Responses: This dimension assesses the LLM's ability to maintain consistency within its generated outputs, ensuring a logical and coherent flow of information.
\end{itemize}

Our proposed benchmark dataset surpasses the limitations of existing approaches by incorporating a simple and feasible automated evaluation method. This innovative approach presents a significant leap forward in the quest to ensure the trustworthiness of LLMs by providing a robust and efficient method for detecting and mitigating hallucinations.


\section{Related Works}
Over the past year, several works have investigated the cause, effect, and detection of hallucinations of different LLMs. Most of the work has been focused on hallucination from the perspective of the factuality of the response and faithfulness to the prompt. Some benchmark datasets have been proposed for hallucination detection as well.

The majority of datasets proposed for assessing hallucinations predominantly concentrate on the detection of hallucinated content within the generated output ~\cite{li2023halueval}~\cite{yang2023new}~\cite{zhao2024felm}~\cite{HaluEval-Wild}~\cite{zhang2023sac}. These datasets commonly employ LLMs, such as chatgpt, to deliberately generate hallucinatory responses. Subsequently, these responses are annotated through additional phases with LLMs or human experts. The annotated data is then utilized to evaluate the efficacy of LLMs in detecting hallucinations within these samples. These benchmark datasets primarily deal with large-scale generated responses, such as passages, necessitating human annotators, or LLMs, to assess performance. However, LLM-based assessments may be susceptible to biases, while human judgments are time-consuming and resource-intensive, leading to the creation of smaller datasets.

Several other datasets have been proposed to evaluate LLM performance across various tasks and methodologies for assessing hallucinations within responses. Some employ static prompts for question-answering tasks~\cite{lin2021truthfulqa}~\cite{cheng2023evaluating}, while~\cite{kasai2024realtime} introduced a method for dynamically generating questions based on real-time news events to assess the adaptability of LLM knowledge bases. These datasets typically utilize multiple-choice question (MCQ) formats for evaluation. However, the MCQ format may not adequately gauge hallucination, as it fails to assess the generative capabilities of LLMs. Models may simply guess answers or identify patterns within the provided options rather than truly generating responses.

In contrast, our dataset is specifically designed to elicit the generative capabilities of LLMs while mitigating reliance on human judgment. Compared to existing datasets, ours is at least twice the size, offering a more robust benchmark for evaluating LLM performance in hallucination detection. A summary of the existing works is given in Table~\ref{table:LR}, and detailed information about each is provided in the supplementary materials.

\setlength{\tabcolsep}{4pt}
\begin{table}
\caption{A summary of existing hallucination benchmarks. Evaluation aspect denotes the category of hallucination being assessed. Granularity of a dataset denotes the level of information being labeled.}
\label{table:LR}
\resizebox{\textwidth}{!}{
\begin{tabular}{llllcccllcc}
\hline
\textbf{}& \textbf{}& \textbf{}& \textbf{}& \multicolumn{3}{c}{\textbf{Evaluation   Aspect}}& \textbf{}& \textbf{}& \multicolumn{2}{c}{Task   Type}\\ \cline{5-7} \cline{10-11}
\textbf{Benchmark}& \textbf{Dataset} & \textbf{Language}& \textbf{Size} & \textbf{Factuality} & \textbf{Faithfulness} & \textbf{Consistency} & \textbf{Granularity}& \textbf{Metirc}            & \textbf{Detection} & \textbf{Evaluation} \\ \hline
Truthful QA~\cite{lin2021truthfulqa}& - & English& 817& \ding{51}& & & Answer& LLM judge, Human&& \ding{51}\\\hline
REALTIMEQA~\cite{kasai2024realtime} & -& English& Dynamic & \ding{51}&&& Answer& Acc, F1&& \ding{51}\\ \hline
\multirow{2}{*}{HaluEval~\cite{li2023halueval}} & Task-specific    & \multirow{2}{*}{English} & 30000& \ding{51}& \ding{51}&& Answer& Acc&&\\
                          & General&& 5000& \ding{51}& \ding{51}&& Answer& Acc& \ding{51}&\\ \hline
\multirow{3}{*}{HaluQA~\cite{cheng2023evaluating}}   & Misleading       && 175& \ding{51}&&& \multirow{3}{*}{Answer} & \multirow{3}{*}{LLM judge} && \ding{51}\\
                          & Misleading-hard  & Chinese& 69& \ding{51}&&&&&& \ding{51}\\
                          & Knowledge        && 206& \ding{51}&&&&&& \ding{51}\\ \hline
FELM~\cite{zhao2024felm}                      & -& English& 3948& \ding{51}& \ding{51}&& Response& Balanced acc \& F1&\ding{51}&\\ \hline
\multirow{3}{*}{PHD~\cite{yang2023new}}      & PHD-Low& \multirow{3}{*}{English} & 100& \ding{51}& \ding{51}&&\multirow{3}{*}{Passage}& P, R, F1& \ding{51}&\\
                          & PHD-Medium       &                          & 100                        & \ding{51}                   & \ding{51}                    &                      &                         & P, R, F2                   & \ding{51}                  &                     \\
                          & PHD-High         &                          & 100                        & \ding{51}                   & \ding{51}                     &                      &                  & P, R, F3                   & \ding{51}                  &                     \\ \hline
\multirow{4}{*}{SAC$^3$~\cite{zhang2023sac}}     & Prime Numbers    &                          & 500                        &                     &                       &                      & \multirow{4}{*}{Answer} & \multirow{4}{*}{AUROC}     & \ding{51}                  &                     \\
                          & Seanator Search  &                          & 500                        &                     &                       &                      &                         &                            & \ding{51}                  &                     \\
                          & HotpotQA         & English                  & 250                        & \ding{51}                   &                       & \ding{51}                    &                         &                            & \ding{51}                  &                     \\
                          & NQ-Open          &                          & 250                        &                     &                       &                      &                         &                            &\ding{51}                  &                     \\ \hline
HaluEval-wild~\cite{HaluEval-Wild}             & -                & English                  & 6505                       & \ding{51}                   &                       &                      & Response                &Acc& \ding{51}                  &           \\  \hline
HalluVault~\cite{li2024halluvault}             & -                & English                  & 14000                       & \ding{51}                   &                       &                      & Response                &Structural similarity&&  \ding{51}         \\  \hline
\multirow{2}{*}{\textbf{DefAn (Proposed)}}            & Public               & English                  & 68093& \ding{51}&\ding{51}& \ding{51}& \multirow{2}{*}{Response}                &\multirow{2}{*}{Hallucination Rate}&&  \ding{51}         \\ 
           & Hidden               & English                  & 7485& \ding{51}&\ding{51}& \ding{51}&                 &&&  \ding{51}         \\  \hline

\end{tabular}}
\end{table}

\section{Proposed DefAn Dataset}
The main goal of this paper is to develop a benchmark to evaluate the factual accuracy of the LLMs, as well as their faithfulness to the given prompt. Existing benchmarks mainly concentrate on detecting hallucinations within the response of LLMs. We believe a specific question-answering benchmark is necessary to understand how LLMs hallucinate factual information. Considering this, we have created a dataset that requires precise responses, and we have gathered the responses from the official documents available online. The LLM output gives an understanding of how they hallucinate over specific details and how much of the facts an LLM provides are to be trusted.

\subsection{Dataset Overview}
The proposed dataset contains around 75,000 samples from various domains of knowledge. The target information of these questions is a specific number, a date, a location or a person. The prompts also ask for specific information from the LLMs.

\subsection{Design Basics}

\textbf{Factuality:} The design of our dataset starts by defining \textit{Factuality}. Li et al.~\cite{li2024dawn} defined factuality hallucination by six fine-grained categories. In general, factuality refers to the degree of accuracy and truthfulness of the generated text about real-world facts or events. It covers how faithfully the generated text represents the information provided or the context in which it is generated. Text can vary in factuality, ranging from entirely factual and precise to speculative or fictional. In text generation tasks, ensuring high factuality is crucial, particularly in applications where accuracy and reliability are paramount, such as news reporting, academic writing, or legal documentation. However, factuality can sometimes be challenging, especially when the generated content involves complex reasoning, interpretation, or subjective perspectives. The existing benchmarks mainly focus on claims made in responses generated by LLMs. Even the QA datasets focus primarily on world knowledge. We have collected samples from diverse domains of world knowledge. We have also collected questions from the math domain that test the understanding of mathematics questions and reasoning. These domains serve as tools to comprehend the characteristics of the hallucinated response of the LLMs.

\noindent\textbf{Faithfulness:} A primary objective of the dataset is to assess the faithfulness of responses generated by LLMs to the provided prompts. To achieve this, prompts are carefully crafted to invoke specific answers, facilitating a focused evaluation process. Even if a generated response contains accurate information, a deviation from the prescribed format is considered unfaithful to the prompt. This emphasis on prompt fidelity ensures that the evaluation accurately reflects the LLMs' ability to produce responses that align closely with the intended context and requirements.

\noindent\textbf{Consistency:} One crucial aspect of the dataset evaluation involved examining whether language models consistently generated responses for the same question over time and across paraphrased versions. To achieve this, each sample underwent rigorous testing through 15 paraphrased versions, allowing for a comprehensive assessment of response consistency as shown in Table~\ref{table:paraphrasing}.

\noindent\textbf{Granularity:} The granularity of a dataset refers to the level of detail or specificity at which the data is organized and structured. In text generation tasks, granularity often pertains to the distinction between responses, claims, and segments within the dataset. We strategically design prompts so that the generated response becomes the sole claim, ensuring clarity and precision in the evaluation process. This approach enhances user friendliness and specificity, allowing a more targeted assessment of the generated content against the provided prompts. By carefully considering the granularity of the dataset, we can streamline evaluation procedures and facilitate a more accurate analysis of text generation model performance.

\noindent\textbf{Category:} The dataset has been partitioned into two categories: the public and hidden datasets. The public dataset will be accessible to evaluate the performance of various LLMs and their respective modifications. Conversely, the hidden dataset, possessing a similar structure to the public dataset, will remain private and serve as a benchmark for model performance assessment. This deliberate division ensures that models trained on the benchmark dataset do not exhibit inflated performance metrics solely due to familiarity with the dataset during training, thus safeguarding the integrity of benchmarking evaluations. The privacy of the hidden dataset is essential to maintaining the integrity and validity of benchmarking procedures.

\begin{table}
\begin{center}
\caption{Paraphrasing of questions. Each sample is paraphrased 15 times initially with the help of chatGPT. Human experts annotated later to maintain the accuracy of the prompts.}
\label{table:paraphrasing}
\resizebox{\textwidth}{!}{%
\begin{tabular}{l|l}
\hline
\noalign{\smallskip}
Original prompt & \textit{Which team was the runner up of 2010 FIFA world cup? } \\ \cline{1-2}
& \textit{1. Who was the second-place finisher in the 2010 FIFA World Cup? } \\
& \textit{3. What country came in second in the 2010 FIFA World Cup?} \\
& \textit{4. Which team ended up as the runner-up in the 2010 FIFA World Cup?} \\
& \textit{6. Who clinched the runner-up spot in the 2010 FIFA World Cup?} \\
& \textit{7. Which country was the second-place holder in the 2010 FIFA World Cup?} \\
Paraphrased & \textit{8. Which team secured the second position in the 2010 FIFA World Cup?} \\
& \textit{9. What nation finished as the runner-up in the 2010 FIFA World Cup?} \\
& \textit{10. In the 2010 FIFA World Cup, which team came in second?} \\
& \textit{11. Who ended up as the runner-up in the 2010 FIFA World Cup?} \\
& \textit{12. Which country attained the runner-up position in the 2010 FIFA World Cup?} \\
& \textit{13. Who was the second-best team in the 2010 FIFA World Cup?} \\
& \textit{14. Which nation was the runner-up in the 2010 FIFA World Cup?} \\
& \textit{15. What team took second place in the 2010 FIFA World Cup?} \\
\hline
\end{tabular}}
\end{center}
\end{table}

\subsection{Factuality Domains}
\begin{figure}[htp]
    \centering
    \includegraphics[width= 0.5\columnwidth]{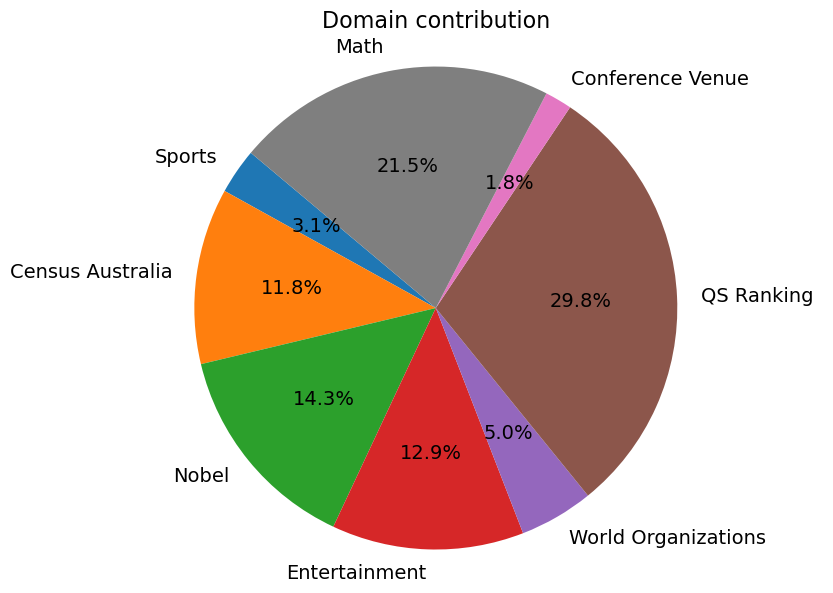}
    \caption{Distribution of prompts by domain}
    \label{fig:domain}
\end{figure}

The proposed dataset contains questions from eight domains of word knowledge and mathematical problems with logical reasoning. They are- Sports, Census Australia, Nobel, Entertainment, World organizations, QS ranking, Conference Venue and Math. Among these, the Sports domain contains information about FIFA World Cup finals\footnote{https://www.rsssf.org/tablesw/worldcup.html}. Census Australia\footnote{https://www.abs.gov.au/census/find-census-data/quickstats/2021/1} archives the statistical information from the Australian Bureau of Statistics census from 2001 to 2021. The Nobel domain contains information about all Nobel laureates\footnote{https://www.nobelprize.org/prizes/lists/all-nobel-prizes/} for different categories. The entertainment domain comprises winners' information and their birthdates for OSCAR winners\footnote{https://awardsdatabase.oscars.org/}. The joining date for the member states of the United Nations (UN)\footnote{https://www.un.org/en/about-us/member-states } and Organization for Islamic Cooperation (OIC)\footnote{https://www.oic-oci.org/states/?lan=en} is archived in word organization. In QS ranking\footnote{https://www.qs.com/reports-whitepapers/qs-world-university-rankings-2024-results-table-excel/}, we accumulate the ranking information for educational institutions. The host location for top conferences is gathered for the Conference venue. In Math\footnote{https://github.com/google-deepmind/AQuA}, the domain includes problems comprising math-related questions designed to assess LLMs' algebraic proficiency and reasoning abilities. Table~\ref{table:dataset_overview} shows an overview of the domains, while Figure~\ref{fig:domain} depicts the distribution of the prompts.  

\subsection{Question Generation}
Generating samples for a QA dataset is a long process that involves several steps to ensure the data's quality, reliability, and consistency. Initially, we gathered information from various official sources such as government publications, academic papers, and official websites. This diverse pool of sources guarantees that the data collected is comprehensive, accurate, and up-to-date. Importantly, each piece of information is carefully examined to ensure its relevance and authenticity, with an emphasis on publicly available content to maintain transparency and accessibility.

Once the information is compiled, clear and specific questions and queries are formulated to extract targeted knowledge from the dataset. These questions are designed to be unambiguous, prompting for particular details or facts directly supported by the collected information. The goal is to create a set of questions that cover a wide range of topics and require precise answers.

To further evaluate the LLMs, each question is paraphrased multiple times to assess the consistency of responses generated by language models. This iterative process helps identify potential inconsistencies or ambiguities in the dataset, ensuring that the LLMs produce coherent and accurate answers across variations of the same question. We use ChatGPT to generate initial samples to paraphrase the questions. The human experts checked these samples to ensure the prompt adhered to the original meaning and invoked the same response. A sample question paraphrasing is shown in Table~\ref{table:paraphrasing}.

\setlength{\tabcolsep}{5pt}
\begin{table}
\begin{center}
\caption{Overview of the domains of proposed dataset. Response type denotes the type of the answers in the datasets. The column \textit{Paraphrased} indicates whether the samples in that domain are paraphrased or not.}
\label{table:dataset_overview}
\begin{tabular}{lccllllc}
\hline\noalign{\smallskip}
& \multicolumn{2}{c}{\textbf{\# of samples}} & \multicolumn{4}{c}{\textit{Response type}} &  \\\cline{2-3}\cline{4-7}
\textbf{Domains} & \textbf{Public}&\textbf{Hidden} & \textit{}{Date} & \textit{Numeric} & \textit{Name} & \textit{Location} & \textbf{Paraphrased}  \\
\noalign{\smallskip}
\hline
\noalign{\smallskip}
Sports & 1305 & 1005 & \ding{51}& \ding{51}&\ding{51} & \ding{51}& \ding{51} \\ 
Census Australia & 7905 & 1005 &  &\ding{51} & & & \ding{51} \\ 
Nobel Prize & 9795 & 1005 & & &\ding{51} & & \ding{51} \\
Entertainment & 8715 & 1005 & \ding{51}& &\ding{51} & & \ding{51} \\
World Organizations & 2745 & 1005 & \ding{51} & & & & \ding{51} \\
QS Ranking & 21495 & 1005 & & \ding{51}& && \ding{51} \\
Conference Venue & 915 & 450 && & &\ding{51}& \ding{51} \\
Math & 15218 & 1005 & & \ding{51}& &&  \\
\hline
\end{tabular}%
\end{center}
\end{table}
\setlength{\tabcolsep}{1.4pt}

\section{Experiment}
Our experiment evaluates the hallucination of publicly available LLMs, analyzing their performance in terms of factuality, faithfulness, and consistency, and identifies potential use cases for our dataset.

\subsection{Experimental Setup}
\noindent{\textbf{LLMs under the scrutiny:}} In our study, we utilized both open-source and closed-source LLMs to evaluate their performance on our dataset. The models employed include zephyr~\cite{zephyr_alignment_handbook2023}, mixtral-8x70b~\cite{jiang2024mixtral}, GPT-3.5~\cite{gpt35}, LLaMA 2~\cite{touvron2023llama}, LLaMA 3~\cite{llama3}, and Gemini Pro~\cite{gemini}. These models represent diverse architectures and capabilities, providing a comprehensive overview of LLM performance across different platforms.

GPT-3.5, developed by OpenAI, is a closed-source model known for its robust language understanding and generation capabilities. LLaMA~2 and LLaMA~3 are open-source models, offering transparency and the ability to fine-tune the models to specific tasks, which is advantageous for research and development purposes. Gemini Pro, a proprietary model, was also included to compare the performance of enterprise-level solutions. We accessed GPT-3.5 and LLaMA~2 using the OpenAI API, facilitating seamless integration and testing of the model within our workflow. For Gemini Pro, we leveraged Google Cloud Services to manage these models.

\begin{table}
\begin{center}
\caption{An overview of all the models used for evaluation. The parameters correspond to the model we used. The context window denotes the maximum allocated context window for the model used. Accessibility is the platform used to access these models.}
\label{table:llm_models}
\begin{tabular}{llccl}
\hline
\textbf{LLMs}  & \textbf{Developer} & \textbf{Parameters} & \textbf{Context Window} & \textbf{Accessibility} \\ \hline
GPT 3.5~\cite{gpt35}& OpenAI& 175 B& 4 K& OpenAI API\\
Llama 2~\cite{touvron2023llama}& Meta& 7 B& 4 K& Llama API\\
Llama 3~\cite{llama3}& Meta& 8 B& 8 K& Lemonfox API\\
Gemini 1.0 pro~\cite{gemini} & Google DeepMind& Unrevealed& 33 K& Google Gemini API\\
mixtral-8x7b~\cite{jiang2024mixtral}& Mistral AI& 7 B& 8 K& Lemonfox API\\
zephyr 7B Beta~\cite{zephyr_alignment_handbook2023} & Mistral AI& 7 B& 8 K& Lemonfox API\\ \hline          
\end{tabular}

\end{center}
\vspace*{-0.7cm}
\end{table}

\noindent{\textbf{Metrics:}} We evaluate the performance of the models based on three perspectives: factual accuracy, faithfulness to prompts, and consistency with paraphrased prompts. Each of these requires a separate metric for evaluation. Let's assume that we have a total $n$ number of questions in the dataset, and among them, $k$ is unique. Others include the paraphrased versions of them. For every question, $q_i$, a response $r_i$ is generated from the LLM.

For the evaluation of FCH, we propose using the FCH rate, which denotes the percentage of the response with the hallucinated fact. FCH rate can be calculated as $\frac{\sum_{i=1}^{n}C_i}{n}$ where $C_i$ is 1 if $r_i$ is incorrect and 0 otherwise.

To measure the Prompt Misalignment Hallucination (PMH), we propose to use PMH rate, calculated as $\frac{\sum_{i=1}^{n}f_i}{n}$, where $f_i$ is 1 if $r_i$ contain PMH and 0 otherwise. 

For measuring consistency, we used Response Consistency (RC), calculated as follows: $\text{RC} = \frac{\sum_{i=1}^{n} \text{Consistency}_i}{n}$. Consistency denotes the percentage of responses that have the same claim.



\section{Result Analysis}
The results from the experiment reveal the hallucination rates of six language models—Zephyr, Mixtral, Llama3, Llama2, GPT-3.5, and Gemini—across eight domains.

\subsection{Performance comparison for specific domains}
This section presents the domain-wise performance of each LLM model. For each domain, we have two sections- public and hidden. 

\noindent\textbf{FCH rate.} Each model's performance was assessed based on the correctness of the factual claim. A bigger value in FCH denoting more hallucination indicates that the model is less trustworthy for factual claims. The FCH rate in each domain is presented in Table~\ref{table:result_fch}.

Domains that require specific numeric information or dates, such as Census, QS Ranking, and Math, exhibit more severe hallucination rates in both public and hidden datasets. This suggests that models struggle significantly with generating accurate numbers. For instance, all models display perfect scores of 1 in the Census domain, indicating a high rate of generating incorrect numbers. High scores in QS Ranking and Math indicate significant challenges in maintaining accuracy with numeric data.

Conversely, domains like Sports, Entertainment, and World Organizations, which typically require names and locations, face less severe hallucinations. Zephyr, for example, shows relatively lower hallucination rates in these domains, with scores improving from 0.50 to 0.29 in Sports and from 0.68 to 0.20 in Entertainment when transitioning from the public to the hidden dataset. This pattern suggests that LLMs perform better when generating non-numeric responses.

Among the models, performance varies considerably across domains and dataset types (hidden vs. public). Overall, Gemini demonstrates the best performance, consistently achieving lower hallucination rates, particularly in domains requiring names and locations. Conversely, Zephyr performs the worst across most domains, especially those requiring specific numeric responses. The other models, such as Llama3, Llama2, and GPT-3.5, exhibit moderate performance with significant variability depending on the domain and dataset type. Notably, while Llama2 and Llama3 perform better in some numeric-focused domains, they still struggle with maintaining accuracy in responses involving specific numbers.

\setlength{\tabcolsep}{5pt}
\begin{table}
\begin{center}
\caption{FCH rate for specific domain. The best results are in bold and a higher value indicates worse performance.}
\vspace{-2mm}
\label{table:result_fch}
\resizebox{\textwidth}{!}{%
\begin{tabular}{lcccccccccccccccc}
\hline
   & \multicolumn{2}{c}{\textbf{Sports}} & \multicolumn{2}{c}{\textbf{Census}} & \multicolumn{2}{c}{\textbf{Nobel}} & \multicolumn{2}{c}{\textbf{Entertainment}} & \multicolumn{2}{c}{\textbf{World Organizations}} & \multicolumn{2}{c}{\textbf{QS Ranking}} & \multicolumn{2}{c}{\textbf{Conf. Venue}} & \multicolumn{2}{c}{\textbf{Math}} \\ \cline{2-17}
   & \textit{Public} & \textit{Hidden} & \textit{Public}  & \textit{Hidden}  & \textit{Public}  & \textit{Hidden} & \textit{Public}  & \textit{Hidden} & \textit{Public}    & \textit{Hidden}   & \textit{Public}    & \textit{Hidden}    & \textit{Public}     & \textit{Hidden}    & \textit{Public} & \textit{Hidden} \\ \hline
\textbf{zephyr}  & 0.50 & 0.29 & 1.00 & 1.00 & 0.91 & 0.93 & 0.68 & 0.20 & 0.95 & 0.92 & 0.94 & 0.98 & 0.82 & 0.95 & 0.99 & 0.99 \\
\textbf{mixtral} & 0.20 & 0.13 & 1.00 & 1.00 & 0.59 & 0.60 & 0.56 & \textbf{0.11} & 0.69 & 0.44 & \textbf{0.88} & 0.98 & 0.52  & 0.63 & \textbf{0.98} &\textbf{ 0.97} \\
\textbf{llama3}  & 0.44 & 0.30 & 1.00 & 1.00 & 0.63 & 0.70 & 0.29 & 0.19 & 0.71 & 0.73 & 0.97 & 0.99 & 0.65  & 0.87 & 1.00 & 0.99 \\
\textbf{llama2}  & \textbf{0.15} & \textbf{0.09} & 1.00  & 1.00  & 0.90  & 0.90 & 0.33  & 0.17 & 0.85 & 0.74   & 0.93 & 0.99 & 0.85  & 0.88 & \textbf{0.98} & 0.98 \\
\textbf{gpt 3.5} & 0.17 & 0.11 & 1.00  & 1.00  & \textbf{0.35}  & \textbf{0.52} & \textbf{0.10}  & 0.19 & 0.57 & 0.38   & 0.93 & 0.98 & \textbf{0.31}  & 0.60 & \textbf{0.98} & 0.98 \\
\textbf{gemini}  & 0.21 & \textbf{0.09} & 1.00  & 1.00  & \textbf{0.35}  & \textbf{0.52} & 0.42  & 0.14 & \textbf{0.54} & \textbf{0.31}   & 0.97 &\textbf{ 0.96 }& 0.47  & \textbf{0.51} & 0.99 & 0.99  \\ \hline        
\end{tabular}
}

\vspace*{-0.5cm}
\end{center}
\end{table}
\setlength{\tabcolsep}{1.4pt}

\noindent\textbf{PMH rate:} Here, prompt misalignment refers to the degree to which a response accurately deviates from the prompt. It may deviate by generating long passages of text instead of giving definitive answers, or it may give totally out-of-context information or provide information in the wrong format.

The data in Table~\ref{table:result_pmh} reveals that prompt misalignment is predominantly model-specific rather than domain-specific. Most models exhibit misalignment issues across all domains, indicating a general challenge in generating responses that accurately align with the given prompts. However, certain models demonstrate a higher adherence to prompts compared to others.

Zephyr and Mixtral show the highest rates of prompt misalignment across all domains, with values close to or at 1.00 in most cases, indicating a significant difficulty in producing responses that match the prompt. For instance, Zephyr's misalignment rates in Sports and Census are exceptionally high, with public dataset values of 0.87 and 1.00, respectively.

In contrast, models like Gemini and Llama3 perform considerably better at maintaining prompt alignment. Gemini, for example, exhibits very low misalignment rates, with values such as 0.01 in Census and 0.04 in Math for the public dataset and similarly low rates in the hidden dataset. Llama3 also shows lower misalignment rates in several domains, such as a public dataset rate of 0.18 in Sports and 0.01 in Entertainment, although it struggles more in domains like Census.

\setlength{\tabcolsep}{5pt}
\begin{table}
\begin{center}
\caption{PMH rate for specific domain. The best results are in bold and a higher value indicates worse performance. }
\vspace{-2mm}
\label{table:result_pmh}
\resizebox{\textwidth}{!}{%
\begin{tabular}{l cc cc cc cc cc cc cc cc}
\hline
   & \multicolumn{2}{c}{\textbf{Sports}} & \multicolumn{2}{c}{\textbf{Census}} & \multicolumn{2}{c}{\textbf{Nobel}} & \multicolumn{2}{c}{\textbf{Entertainment}} & \multicolumn{2}{c}{\textbf{World Organizations}} & \multicolumn{2}{c}{\textbf{QS Ranking}} & \multicolumn{2}{c}{\textbf{Conf. Venue}} & \multicolumn{2}{c}{\textbf{Math}} \\ \cline{2-17}
   & \textit{Public} & \textit{Hidden} & \textit{Public}  & \textit{Hidden}  & \textit{Public}  & \textit{Hidden} & \textit{Public}  & \textit{Hidden} & \textit{Public}    & \textit{Hidden}   & \textit{Public}    & \textit{Hidden}    & \textit{Public}     & \textit{Hidden}    & \textit{Public} & \textit{Hidden} \\ \hline
\textbf{zephyr}  & 0.87 & 0.98 & 1.00 & 1.00 & 0.96 & 0.98 & 0.76 & 0.41 & 0.99 & 0.99 & 1.00 & 1.00 & 1.00 & 1.00 & 1.00 & 1.00 \\
\textbf{mixtral} & 0.95 & 0.89 & 1.00 & 1.00 & 0.94 & 0.99 & 0.87 & 0.71 & 1.00 & 1.00 & 1.00 & 1.00 & 0.97 & 0.99 & 0.98 & 0.98 \\
\textbf{llama3}  & 0.18 & 0.34 & 0.98 & 0.99 & 0.16 & \textbf{0.26} & \textbf{0.01} & 0.03 & 0.78 & 0.74 & 0.52 & 0.56 & 0.24 & 0.26 & 0.04 & 0.04 \\
\textbf{llama2}  & 0.07 & 0.09 & 0.96 & 0.99 & 0.48 & 0.85 & 0.04 & \textbf{0.01} & 0.74 & \textbf{0.72} & 1.00 & 0.99 & 0.64 & 0.57 & 0.02 & \textbf{0.01 }\\
\textbf{gpt 3.5} & 0.17 & 0.16 & 0.55 & 0.49 & 0.14 & 0.41 & 0.31 & 0.33 & 0.75 & 0.88 & 0.55 & 0.62 &\textbf{ 0.17} & 0.22 & 0.38 & 0.36 \\
\textbf{gemini}  & \textbf{0.06} & \textbf{0.05} & \textbf{0.01 }& \textbf{0.00} & \textbf{0.12} & 0.36 & 0.06 & \textbf{0.01} & \textbf{0.57} & 0.80 & \textbf{0.04} & \textbf{0.00 }& 0.27 & \textbf{0.20} & \textbf{0.01} & 0.02 \\ \hline        
\end{tabular}
}
\end{center}
\end{table}
\setlength{\tabcolsep}{1.4pt}

\noindent\textbf{RC:}
RC measures the prowess to generate consistent responses over paraphrased versions of the same prompt. The bigger the value, the better the performance. RC is measured for all the domains except the math domain, as the prompts in this domain are not paraphrased. The data is shown in Table~\ref{table:result_rc}.

The data shows that models generally exhibit more significant inconsistency when generating specific numbers, as seen in domains like census, while other domains tend to elicit more consistent responses over paraphrased prompts. Models like Gemini, LLaMA~2, LLaMA~3 and GPT show more consistency for the domains other than census and QS ranking. Other models are inconsistent.

\begin{table}
\caption{RC score for specific domain. The best results are in bold and the  higher value denotes better performance.}
\label{table:result_rc}
\resizebox{\textwidth}{!}{%
\begin{tabular}{l cc cc cc cc cc cc cc}
\hline
 & \multicolumn{2}{c}{\textbf{Sports}} & \multicolumn{2}{c}{\textbf{Census}} & \multicolumn{2}{c}{\textbf{Nobel}} & \multicolumn{2}{c}{\textbf{Entertainment}} & \multicolumn{2}{c}{\textbf{World Organizations}} & \multicolumn{2}{c}{\textbf{QS Ranking}} & \multicolumn{2}{c}{\textbf{Conf. Venue}}\\ \cline{2-15}
 & \textit{Public} & \textit{Hidden} & \textit{Public} & \textit{Hidden} & \textit{Public} & \textit{Hidden} & \textit{Public} & \textit{Hidden} & \textit{Public} & \textit{Hidden} & \textit{Public} & \textit{Hidden} & \textit{Public} & \textit{Hidden} \\ \hline
\textbf{zephyr} & 0.19 & 0.15 & 0.07 & 0.07 & 0.10 & 0.11 & 0.43 & 0.59 & 0.13 & 0.15 & 0.13 & 0.10 & 0.47 & 0.43 \\
\textbf{mixtral} & 0.19 & 0.28 & 0.07 & 0.07 & 0.12 & 0.09 & 0.38 & 0.26 & 0.13 & 0.22 & 0.07 & 0.07 & 0.78 & 0.74 \\
\textbf{llama3} & 0.60 & 0.62 & 0.07 & 0.07 & 0.46 & 0.52 & 0.81 & 0.84 & 0.50 & 0.46 & 0.11 & 0.08 & 0.58 & 0.50 \\
\textbf{llama2} & \textbf{0.94} & \textbf{0.97} & 0.07 & 0.07 & 0.36 & 0.21 & \textbf{0.96} & 0.97 & 0.28 & 0.31 & 0.09 & 0.07 & 0.47 & 0.43 \\
\textbf{gpt 3.5} & 0.77 & 0.86 & 0.07 & 0.07 & \textbf{0.80} & 0.62 & 0.67 & 0.66 & 0.28 & 0.23 & \textbf{0.21} & 0.15 & \textbf{0.84} & 0.73 \\
\textbf{gemini} & 0.82 & 0.91 & 0.07 & 0.07 & 0.79 & \textbf{0.74} & 0.89 & \textbf{0.99} & \textbf{0.79} & \textbf{0.82} & 0.15 & \textbf{0.16} & 0.78 & \textbf{0.76} \\ \hline
\end{tabular}
}

\vspace*{-.7cm}
\end{table}

\subsection{Overall Performance}
\begin{figure}[tp]
\centering
\begin{tabular}{cc}
\includegraphics[width= .49 \linewidth]{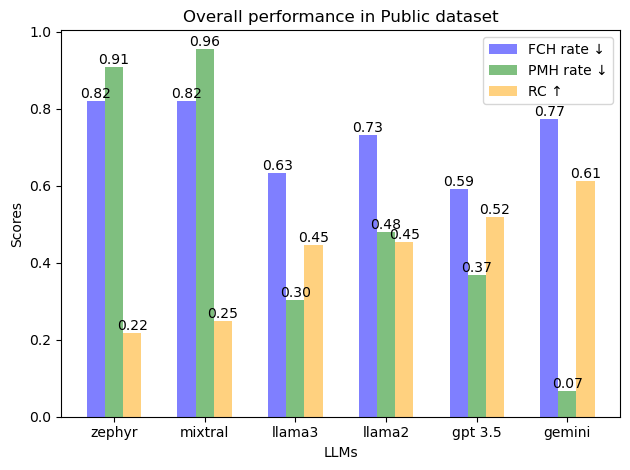}&
\includegraphics[width= .49 \linewidth]{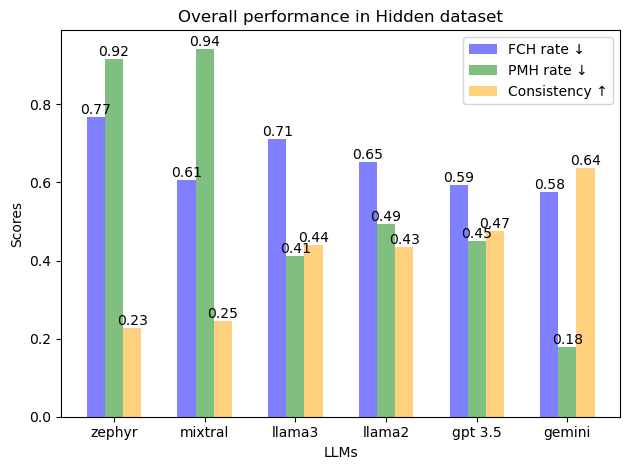}\\
a) & b)\\
  
\end{tabular}
\caption{The performance comparison on all three evaluation metrics for LLMs in a) public and b) hidden datasets.}
\label{fig:res_all} 
\vspace{-4mm}
\end{figure}

Figure~\ref{fig:res_all} illustrate the performance of the LLMs based on the three metrics we proposed. The model performance analysis reveals noteworthy trends across various evaluation metrics. First, focusing on factual correctness (FCH), it becomes evident that most models face challenges in generating factually accurate responses. Both Llama~2 and GPT-3.5 exhibit a moderate level of performance across both public and hidden datasets, suggesting a better ability to produce factually correct responses. However, models such as Llama~3, Gemini, and Mixtral display fluctuating performance, indicating variability in their accuracy in generating factually correct responses across different datasets.

Considering PMH, certain models, notably Zephyr and Mixtral, demonstrate severe deviations from the provided prompts. This suggests significant challenges in accurately adhering to the provided prompts. Conversely, Gemini emerges as a standout performer in both datasets, showcasing its superior adherence capability. Other models exhibit moderate performance, with varying degrees of deviation from the provided prompts.

Lastly, analyzing response consistency, Gemini stands out as the most consistent model across paraphrased prompts. Its ability to maintain coherence and consistency across different variations of the prompts. Models like Mixtral and Zephyr demonstrate the worst performance, suggesting difficulties producing coherent responses across paraphrased prompts. Other models exhibit moderate levels of performance in response consistency.

    

\section{Limitation \& Future Work}
Despite the robustness and utility of the proposed benchmark dataset, there are several limitations that may be addressed in future work.

\noindent\textbf{Limited coverage of knowledge domain:} The dataset currently covers a few knowledge domains. To enhance the comprehensiveness of the benchmark, one of the ways is to include information from additional domains. Future dataset versions may incorporate domains such as science and technology, medicine, economy, and ethics. This expansion will provide a more holistic evaluation of LLMs across various topics. However, the inclusion of these specialized domains presents significant challenges, as it requires annotations from domain experts and the careful crafting of prompts to invoke definitive responses from the LLMs.

\noindent\textbf{Incorporation of novel metrics:} Introducing new evaluation metrics is needed to capture more aspects of LLM performance. One such metric that could be valuable is sycophancy~\cite{sharma2023towards}, which assesses the confidence of the generated response. Incorporating this metric would allow for a more nuanced understanding of how LLMs handle uncertain or ambiguous prompts and how confident they are in their responses.

\section{Conclusion}
This paper introduces a comprehensive benchmark dataset designed for evaluating hallucinations in LLMs. To facilitate accurate assessment and evaluation of hallucinations in the generative capabilities of LLMs, the dataset ensures that target responses have definitive answers. The resulting dataset combines responses and claims, enhancing its granularity. Comprising over 75,000 prompts across nine distinct domains, the dataset features target answers in the form of names, places, dates, or specific numeric values. We have proposed three evaluation metrics: factual accuracy, faithfulness accuracy, and consistency accuracy. Utilizing our dataset, we tested several prominent public LLMs, including GPT-3.5, LLaMA 2.0, LLaMA 3.0, Gemini 1.0 Pro, Claude, Mistral, and Zephyr. Our findings reveal that most LLMs exhibit hallucinations, both factually and in terms of faithfulness to the prompt. For consistency, apart from specific numeric values, most LLMs were consistent in their responses to paraphrased prompts. Overall, performance in generating names, places, and dates was moderate, but significant hallucinations occurred when numeric values were required. In summary, our dataset is comprehensive, challenging, and easy to assess, making it a valuable benchmark for evaluation.

\section*{Supplementary Materials}
\begin{appendices}

In the supplementary, we initially present information regarding the knowledge domains of our dataset. This is followed by the methodology employed to gather the dataset. Finally, we furnish the details regarding the evaluation.

\section{Knowledge Domains}
To construct the knowledge base, we gathered information from eight domains, ranging from sports and entertainment to world politics. These domains are
\begin{itemize}
    \item \textbf{Sports:} The FIFA World Cup, organized by the Fédération Internationale de Football Association (FIFA), is the premier international soccer tournament held every four years. The inaugural World Cup occurred in 1930 in Uruguay, with the host nation securing the first championship title. Over the decades, the tournament has expanded in scope and influence, now featuring 32 teams in its final stages\footnote{https://www.rsssf.org/tablesw/worldcup.html}. In this domain, we have generated information about the FIFA World Cup finals from 1930 to 2022. The target information ranges from all the domains, host stadium and city (Location), winner/runner-up (Country), and attendance (Numeric).
    
    \item \textbf{Census Australia:} The Australian Census, conducted by the Australian Bureau of Statistics (ABS) every five years, is a comprehensive survey that collects detailed information about the country's population and housing. It provides essential data on demographics, socioeconomic status, and living conditions, which are crucial for government planning and policy-making. The most recent Census was held in 2021\footnote{https://www.abs.gov.au/census/find-census-data/quickstats/2021/1}, capturing a snapshot of Australia's diverse and evolving society. This domain contains only numeric information. We obtained the age group-specific population from the ABS report. This domain contains around 9000 questions regarding the population of different regions of Australia in a specific year.
    
    \item \textbf{Nobel Prize:} The Nobel Prize is one of the most prestigious awards in the world, honoring individuals and organizations for outstanding contributions in the fields of physics, chemistry, medicine, literature, peace, and economic sciences. Established by the will of Alfred Nobel, the inventor of dynamite, the prizes have been awarded annually since 1901, recognizing advancements that have had a significant impact on humanity. Recipients of the Nobel Prize often represent the pinnacle of achievement in their respective fields, inspiring generations and shaping the course of history. This domain contains questions about the winner of the Nobel Prize every year. The information is collected from the official website of the Nobel Prize organization.\footnote{https://www.nobelprize.org/prizes/lists/all-nobel-prizes/}
    
    \item \textbf{Entertainment:} The Oscars, formally known as the Academy Awards, celebrate excellence in the film industry, recognizing outstanding achievements in various categories such as Best Picture, Best Actor, and Best Director. Held annually by the Academy of Motion Picture Arts and Sciences since 1929, the Oscars are a highlight of the entertainment calendar, showcasing the talent and creativity of filmmakers from around the world\footnote{https://awardsdatabase.oscars.org/}. In the entertainment domain, prompts are designed to invoke the names of the winners of various Oscar categories, including best actor, best director, and best film, among others. It also includes the birthdates of the winners and the titles of the films for which they were awarded.
    
    \item \textbf{World organizations:} This domain covers two prominent world organizations: the United Nations (UN) and the Organization of Islamic Cooperation (OIC). The UN is an international organization founded in 1945, tasked with maintaining international peace and security, promoting sustainable development, and upholding human rights. Comprising 193 member states, the UN serves as a forum for diplomacy, negotiation, and cooperation on global issues ranging from climate change to humanitarian crises. In this domain, we designed questions about the date of joining for each member states\footnote{https://www.un.org/en/about-us/member-states}. The Organization of Islamic Cooperation (OIC) is the second-largest intergovernmental organization after the United Nations, representing 57 member states with significant Muslim populations. Established in 1969, the OIC aims to safeguard the interests of Muslims worldwide, promote solidarity among member states, and foster cooperation in economic, social, and cultural spheres. Through its collective efforts, the OIC addresses issues ranging from conflict resolution to development, advocating for the global rights and well-being of Muslim communities. The questions generated for this topic are about the joining year of each member states\footnote{https://www.oic-oci.org/states/?lan=en}.
    
    \item \textbf{QS Ranking:} QS World University Rankings is an annual publication of university rankings by Quacquarelli Symonds, a British company. It evaluates universities worldwide based on factors such as academic reputation, employer reputation, faculty/student ratio, citations per faculty, international faculty ratio, and international student ratio. Widely regarded as one of the most influential university rankings globally, QS rankings serve as a valuable resource for students, academics, and policymakers in assessing the quality and reputation of higher education institutions. We have taken the QS ranking of the last three years, from 2022 to 2024\footnote{https://www.qs.com/reports-whitepapers/qs-world-university-rankings-2024-results-table-excel/}. The questions ask for the specific ranking of a university/institute.
    
    \item \textbf{Conference Venue:} Conferences such as Empirical Methods in Natural Language Processing (EMNLP)\footnote{https://dblp.org/db/conf/emnlp/index.html}, European Conference on Computer Vision (ECCV)\footnote{https://dblp.org/db/conf/eccv/index.html}, and Conference on Computer Vision and Pattern Recognition (CVPR)\footnote{https://dblp.org/db/conf/cvpr/index.html} are premier events in the fields of natural language processing and computer vision, held annually in various venues worldwide. These conferences are platforms for researchers, academics, and industry professionals to present and discuss the latest advancements, methodologies, and applications in their respective fields. With thousands of attendees from around the globe, these conferences foster collaboration, innovation, and the exchange of ideas, shaping the future of these rapidly evolving disciplines. We are interested in the city that has hosted each conference over the years.
    
    \item \textbf{Math:} We curated a domain comprising math-related questions designed to assess algebraic proficiency and reasoning abilities. With over 16,000 samples sourced from diverse platforms\footnote{https://github.com/google-deepmind/AQuA}\footnote{https://www.kaggle.com/datasets/thedevastator/mathematical-problems-dataset-various-mathematic/} and educational materials, the dataset offers a comprehensive spectrum of mathematical challenges. Ranging from elementary calculations like '1+1' to complex problems like solving differential calculus, the samples encompass a wide range of difficulty levels. Additionally, the dataset incorporates problems necessitating logical reasoning, providing a holistic evaluation of mathematical skills. 
\end{itemize}
Table~\ref{table:domain_sample} contains a summary of the dataset.

\setlength{\tabcolsep}{3pt}
\begin{table}
\begin{center}
\caption{Sample questions from each of the knowledge domains. The column \textit{Target} denotes the expected data type of the answer. Data type \textit{Location} is more specified to \textit{Country} and \textit{City}}
\label{table:domain_sample}
\resizebox{\textwidth}{!}{%
\begin{tabular}{lll}
\hline\noalign{\smallskip}
\textbf{Domain} & \textbf{Sample questions} & Target \\
\noalign{\smallskip}
\hline
\noalign{\smallskip}
Sports & \textit{\enquote{Which team won the 2022 FIFA World Cup?
}} & Country\\\cline{2-3}
& \textit{\enquote{Which team ended up as the runner-up in the 2010 FIFA World Cup?}} & Country \\ \cline{2-3}
& \textit{\enquote{Which stadium hosted the 1994 FIFA World Cup final?}} & Name \\ \cline{2-3}
& \textit{\enquote{What city was the 1962 FIFA World Cup final held in?}} & City \\ \cline{2-3}
& \textit{\enquote{ Which country acted as the host for the 1954 FIFA World Cup final?}} & Country \\ \cline{2-3}
& \textit{\enquote{ What was the number of attendees at the final match of the 2014 FIFA World Cup?}} & Numeric \\ \cline{1-3}
Census Australia & \textit{\enquote{What was the population of Australia in 2001?}} & Numeric \\\cline{2-3}
 & \textit{\enquote{In 2001, what was the population of New South Wales of age group 0 - 4 years?}} & Numeric \\ \cline{1-3}
Nobel Prize & \textit{ \enquote{Who won the Nobel Prize for Physics in 2001?}} & Name\\ \cline{1-3}
Entertainment & \textit{\enquote{What is Mary Pickford's birth date (mm/dd/yyyy)?}} & Date \\ \cline{2-3}
 & \textit{\enquote{In 2017, for which movie did Gary Oldman win the Oscar? give the movie name only.}} & Name\\ \cline{2-3}
 & \textit{\enquote{What film won the Oscar for best picture in 2015?give one name only.}} & Name \\ \cline{2-3}
 & \textit{\enquote{Winner of 2001 OSCAR winner for best actor?}}  & Name \\ \cline{1-3}
 World Organizations & \textit{\enquote{Exact date of KSA becoming a member of UN}} & Date \\  \cline{2-3}
 & \textit{\enquote{In which year did Bangladesh become a member of OIC?}} & Date \\ \cline{1-3}

QS Ranking & \textit{\enquote{In 2024, where did University of Guelph stand in the QS ranking?}} & Rank \\ \cline{1-3}
Conference & \textit{\enquote{Which city hosted 2022 EMNLP?}} & City \\ \cline{1-3}
Math & \textit{\enquote{what is the area of square field whose side of length 13 m?}} & Numeric \\ \cline{2-3}
& \textit{\enquote{$5555 \times 9999 =$ ?}} & Numeric \\
\hline
\end{tabular}}
\end{center}
\end{table}
\setlength{\tabcolsep}{1.4pt}

\section{Data Collection}
We chose official websites and databases mentioned in the previous section that are relevant to each specific knowledge domain as our primary data sources. Employing web scraping techniques, we systematically gathered the necessary information stored in Excel files. Human experts formulated sample questions for each prompt type to ensure clarity and precision, thus minimizing potential ambiguities. Python scripts leveraged the collected data and sample questions to generate a comprehensive set of prompts. We meticulously compiled the finalized questions and are now available in CSV and JSON formats. We finally divided the dataset into two sections—hidden and public—for each domain.

\begin{figure}[tp]
\centering

\includegraphics[width= \linewidth]{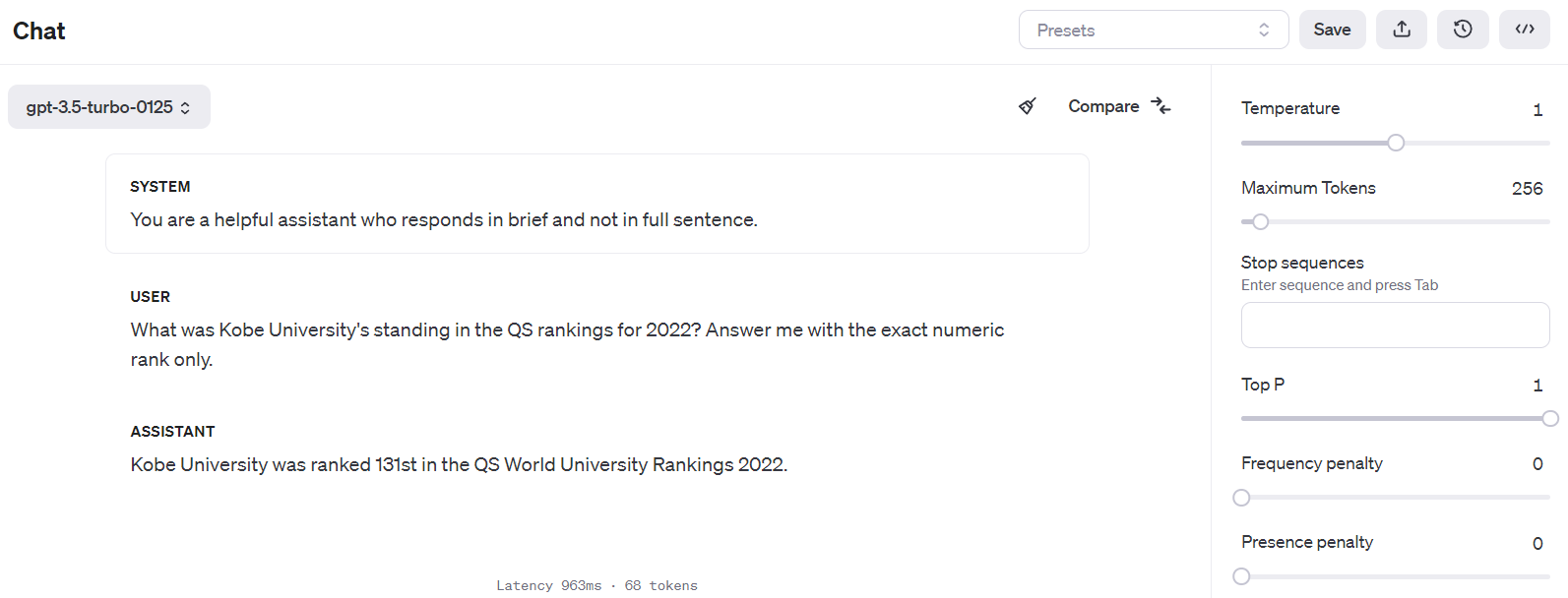}

\caption{Sample prompt execution. Visualized using openAI playground.}
\label{fig:prompts} 
\vspace{-4mm}
\end{figure}

\noindent\textbf{Prompt Execution.}
After preparing the dataset, we generate responses for each selected LLM for analysis using the mentioned prompts. We have used the APIs to access the LLMs. One such example is shown in Figure~\ref{fig:prompts}.
\section{Evaluation}
\noindent\textbf{Claim extraction.}
Upon recording responses from the language models, these responses undergo a rigorous evaluation process. Each response is compared to the reference answer to assess Fact Contradicting Hallucination (FCH) and to the original prompts to evaluate Prompt Misalignment Hallucination (PMH). This involves extracting the factual claims from the responses. Initially, the responses are subjected to basic natural language processing (NLP) pre-processing steps, such as removing punctuation, stopping words, and formatting dates. Subsequently, depending on the target data type, the claims are extracted using a combination of NLP techniques, regular expressions, and string matching. Once the claims are extracted, they are matched against the reference answers for further detailed analysis.

\noindent\textbf{Case study.}
Once the pre-processing is completed, the responses generated by LLM go through the evaluation of FCH, PMH, and RC. Table~\ref{table:evaluation_example} illustrates an example of this evaluation process.

In this example, 15 zero-shot prompts ask for a specific university's QS rank. The responses here are generated by \textit{Gemini 1.0 pro}. Of 15 responses, 3 contain correct answers, making 12 factually incorrect claims. Hence, the FCH rate here is $12/15 = 0.80$.

The prompts are designed to obtain only ranks from the LLMs. 5 out of 15 responses deviate from the instructions provided. The PMH rate here is $5/15 = 0.33$.

To assess response consistency, the maximum frequency of an answer is calculated over the 15 answers. In this example, the most frequent answer is 334, which has a frequency of 4. So, for this set of prompts, the LLM is consistent 4 out of 15 times. The RC value is $4/15 = 0.267$. The final RC value is the average RC for all sets of prompts like table~\ref{table:evaluation_example}.

\begin{table}
\begin{center}
\caption{Response generated by LLMs under evaluation. The cell color denotes PMH and text color denotes the FCH. }
\label{table:evaluation_example}
\resizebox{\textwidth}{!}{%
\begin{tabular}{lcl}\hline
\textbf{prompts} & \textbf{answer} & \textbf{llm-answer} \\ \hline
What was the position of Kobe University in the QS rankings for 2022? Answer me with the exact numeric rank only. & 386 & \cellcolor{YellowGreen} \textcolor{Red}{334} \\
In 2022, where did Kobe University stand in the QS ranking? Answer me with the exact numeric rank only. & 386 & \cellcolor{YellowGreen}\textcolor{Red}{62} \\
What was the QS ranking of Kobe University for 2022? Answer me with the exact numeric rank only. & 386 & \cellcolor{YellowGreen}\textcolor{Red}{334} \\
How did Kobe University fare in the QS rankings for 2022? Answer me with the exact numeric rank only. & 386 & \cellcolor{pink}\textcolor{Red}{Kobe University was not listed in the QS rankings for 2022.} \\
What was Kobe University's QS rank in 2022? Answer me with the exact numeric rank only. & 386 & \cellcolor{YellowGreen} \textcolor{ForestGreen}{386} \\
In 2022, what position did Kobe University secure in the QS rankings? Answer me with the exact numeric rank only. & 386 & \cellcolor{YellowGreen}\textcolor{Red}{195} \\
What was Kobe University's standing in the QS rankings for 2022? Answer me with the exact numeric rank only. & 386 & \cellcolor{pink}Kobe University was ranked \textcolor{Red}{334th} in the QS World University Rankings for 2022. \\
In 2022, what was the QS ranking of Kobe University? Answer me with the exact numeric rank only. & 386 &\cellcolor{YellowGreen} \textcolor{Red}{801 - 1000} \\
Where did Kobe University place in the QS rankings for 2022? Answer me with the exact numeric rank only. & 386 &\cellcolor{pink} \textcolor{Red}{Kobe University is not ranked in the QS rankings for 2022.} \\
What QS ranking did Kobe University achieve in 2022? Answer me with the exact numeric rank only. & 386 & \cellcolor{YellowGreen}\textcolor{Red}{334th} \\
QS ranking of Kobe University in 2022? Answer me with the exact numeric rank only. & 386 & \cellcolor{YellowGreen}\textcolor{Red}{467} \\
What position did Kobe University attain in the QS rankings for 2022? Answer me with the exact numeric rank only. & 386 & \cellcolor{pink} Kobe University was ranked \textcolor{ForestGreen} {386th} in the QS World University Rankings for 2022. \\
What was Kobe University's QS rating for 2022? Answer me with the exact numeric rank only. & 386 & \cellcolor{pink}\textcolor{Red}{Kobe University was not ranked in the QS World University Rankings for 2022.} \\
Give the specific QS rank of Kobe University in 2022 Answer me with the exact numeric rank only. & 386 & \cellcolor{YellowGreen} \textcolor{ForestGreen} {386th} \\
what place did Kobe University achieved in 2022 QS ranking? Answer me with the exact numeric rank only. & 386 & \cellcolor{YellowGreen}\textcolor{Red}{601-650}\\ \hline
\end{tabular}}
\end{center}
\end{table}

\end{appendices}

\bibliographystyle{splncs}
\bibliography{egbib}

\end{document}